\documentclass[11pt]{article}

\usepackage[T1]{fontenc}
\usepackage[utf8]{inputenc}
\usepackage{lmodern}
\usepackage[margin=1in]{geometry}
\usepackage{microtype}
\usepackage{graphicx}
\usepackage{booktabs}
\usepackage{array}
\usepackage{enumitem}
\usepackage{xurl}
\usepackage{hyperref}
\usepackage{caption}
\usepackage{amsmath}
\usepackage{amssymb}

\usepackage{authblk}

\hypersetup{
  colorlinks=true,
  linkcolor=blue,
  citecolor=blue,
  urlcolor=blue,
  pdftitle={Structured Decomposition for Reliable LLM-Generated Access Control Policies},
  pdfauthor={Vatsal Gupta and Darshan Sreenivasamurthy}
}

\setlist[itemize]{leftmargin=1.5em}
\title{
  Structured Decomposition for Reliable\\
  LLM-Generated Access Control Policies
}

\author[1]{
  Vatsal Gupta%
  \thanks{Corresponding author. 
  \href{https://orcid.org/0009-0007-0732-9895}
  {ORCID: 0009-0007-0732-9895}}
}

\author[1]{
  Darshan Sreenivasamurthy%
  \thanks{
  \href{https://orcid.org/0009-0005-3706-546X}
  {ORCID: 0009-0005-3706-546X}}
}

\affil[1]{
  Apple Inc., Cupertino, CA, USA\\
  \texttt{\{vatsal\_gupta,dtumkursreenivas\}@apple.com}
}

\date{}
\begin{document}
\maketitle

\begin{abstract}
The paper presents an LLM-based system that translates natural-language access control policies (NLACPs) into executable Rego code for Open Policy Agent (OPA). It provides a modular, end-to-end pipeline performing policy detection, component extraction, schema validation, linting, compilation, and automated test generation and execution. The system is designed to bridge the gap between human-readable access requirements and machine-enforceable policy-as-code (PaC), with a focus on deployment reliability and security correctness.

We evaluate the system on 372 ACRE-complete access control statements (with non-null subject, action, and resource annotations) against a direct single-prompt LLM baseline to isolate the contribution of structured decomposition and schema-aware validation. The system achieves a 50.3\% end-to-end policy correctness rate compared to 15.3\% for the baseline, representing a 3.3$\times$ improvement. A policy is counted as correct only if it satisfies compilation, linting, and both positive and negative tests, making this a strict measure of deployable correctness.

On security-critical patterns, the system generates correct deny semantics for 87.5\% of deny policies (baseline: 37.5\%), ownership conditions for 100\% of ownership-qualified policies (baseline: 40\%), and status-qualified conditions for 100\% of status-qualified policies (baseline: 55.6\%). These results indicate that structured decomposition and schema-aware validation play a critical role in improving the reliability of LLM-generated authorization policies.
\end{abstract}


\section{Introduction}

Access control is foundational to security and has become increasingly critical with rapid digital transformation and the emergence of autonomous and distributed systems. Traditional approaches that embed authorization logic directly within applications are difficult to audit, maintain, and scale, particularly in dynamic environments that require frequent policy updates and contextual decision-making \cite{ref2}. As a result, organizations are increasingly adopting externalized authorization models, including policy-as-code (PaC) frameworks and Zero Trust architectures, which rely on centralized policy evaluation and continuous verification \cite{ref14,ref9}.

A central challenge in these systems is the translation of human-defined access requirements into machine-enforceable policies. While business stakeholders often express access requirements as natural-language statements, implementing these policies requires expertise in formal policy languages such as XACML \cite{ref21}, NGAC \cite{ref9}, Rego \cite{ref22}, or Cedar \cite{ref8}. This gap between human intent and executable policy logic has motivated research into automated translation of natural-language access control policies (NLACPs).

Recent advances in large language models (LLMs) have accelerated progress in this area. Prior work has explored using LLMs and related techniques to extract policy elements and generate formal representations \cite{ref31,ref15,ref12,ref25,ref7}. While these approaches improve extraction and generation capabilities, they introduce new challenges. LLMs may hallucinate entities, omit critical conditions, or generate syntactically valid but semantically incorrect policies \cite{ref4,ref31}. Furthermore, direct prompting approaches may fail to produce outputs for complex or ambiguous inputs, effectively leaving valid access control requirements unimplemented.

These limitations highlight a key gap in existing work: while prior research focuses on improving extraction accuracy or generation quality, relatively little attention has been paid to input validation, semantic completeness, and deployability of generated policies. In practice, these factors are critical in security-sensitive systems, where missing constraints or incorrect deny semantics can lead to over-permissive access or unintended privilege escalation.

In this paper, we present a structured pipeline for translating NLACPs into executable Rego policies. Instead of relying on direct LLM-based generation, the system decomposes policies into structured components, validates them against schema constraints, and applies multi-stage verification through linting, compilation, and automated testing. This design introduces explicit control points that reduce common failure modes in LLM-based policy generation and improve deployment reliability.

We evaluate the system on 372 ACRE-complete access control statements using the dataset's structured annotations as external ground truth. Compared to a direct single-prompt LLM baseline, our approach significantly improves effective policy coverage and reduces security-critical errors, including missing deny rules and incomplete condition handling.

These results suggest that the primary challenge in LLM-based policy generation is not code synthesis, but reliable interpretation and validation of natural-language inputs. Structured approaches that combine decomposition, schema-aware constraints, and verification are therefore essential for making LLM-generated authorization policies safe and deployable in real-world systems.

Our notion of correctness reflects operational validity under compilation, linting, and test-based verification rather than full semantic equivalence to the original policy intent. A key design element is schema-aware generation, which constrains both extraction and code synthesis to organization-defined attribute vocabularies, preventing hallucinated policy elements and improving consistency.

\subsection{Contributions}

This paper makes the following contributions:

\begin{itemize}
  \item We present the system, a modular pipeline that translates natural-language access control policies into executable Rego policies with integrated validation and testing.
  \item We introduce a structured decomposition of policies into Decision, Subject, Action, Resource, Condition, and Purpose (DSARCP), enabling consistent policy representation, along with three deterministic post-generation mechanisms: redundant-block consolidation, semantic safety guardrails, and condition-aware test repair that operate outside the LLM inference path.
  \item We conduct a systematic controlled comparison of the system against a direct single-shot LLM baseline on 372 ACRE-complete policies, demonstrating a 3.3$\times$ improvement in effective correct Rego coverage (50.3\% vs. 15.3\%) and stronger security semantics across deny-decision, ownership-qualified, and status-qualified policies.
  \item We incorporate schema-aware Rego generation and test-driven verification, including positive-test condition patching and negative-test repair, to improve the deployability and security correctness of generated policies.
  \item We show that structured decomposition and validation significantly improve both policy coverage and generation quality compared to direct prompting approaches, particularly for policies involving conditional and security-sensitive semantics.
\end{itemize}

\section{Background and Related Work}

This section discusses research related to converting NLACPs into executable policies that the system builds upon.

\subsection{Evolution and Policy Engineering}

Narouei et al. \cite{ref18} noted that the manual process of translating NLACPs into formal policies is laborious, expensive, and error prone. As also highlighted by Jayasundara et al. \cite{ref12}, natural language policy documents are inherently ambiguous, resulting in error-prone, inconsistent enforcement when manually converted to formal policy code. Similarly, Yang et al. \cite{ref31} demonstrated that developing policies from high-level organizational requirements is labor intensive and error prone. This led to the use of natural language processing, deep learning, and large language models to automate the task of converting NLACPs into machine-enforceable policies.

Early work in automated access control policy generation, such as Text2Policy proposed by Xiao et al. \cite{ref29} and ACRE from Slankas and Williams \cite{ref26}, relied on rule-based extraction using dependency structures. Later, Narouei and Takabi \cite{ref19} introduced top-down policy engineering frameworks leveraging semantic role labeling and neural models. Abdelgawad et al. \cite{ref1} advanced this with natural-language-to-NGAC graph generation. Recent work such as AutoPAC \cite{ref7} and multi-agent code-orchestrated generation \cite{ref13} extends this trajectory by using LLMs to autonomously generate enforceable policies and infrastructure-as-code configurations.

However, most previous work was not accurate enough to extract the necessary elements like subject, resource, and action from NLACPs \cite{ref18,ref11,ref19,ref20}. Furthermore, prior research \cite{ref18,ref1,ref31,ref15,ref9} did not focus on extracting purpose hidden in these statements, which is a crucial component to support authorization for AI agents \cite{ref28}.

\subsection{LLM-Based Policy Generation}

With advances in large language models, several studies have explored their usage to automate translation \cite{ref10,ref16,ref13,ref32,ref31,ref15} with improved extraction results, but they often lack domain-specific knowledge, which results in extraction of invalid subjects or resources. Some recent works \cite{ref12,ref25,ref7} employ retrieval-augmented generation methods that rely on organization-specific corpora to ground policy synthesis in domain context. While this improves relevance, it can reduce flexibility for large enterprises where individual application teams seek lightweight integration. Such pipelines typically need reindexing when organizational data or terminology changes, and prompt templates may require adaptation or fine-tuning depending on implementation.

Also, most prior research \cite{ref18,ref11,ref19,ref20} reported limited accuracy on component extraction and entity grounding, particularly for complex or multi-clause policies. Moreover, as Chen et al. \cite{ref4} and Yang et al. \cite{ref31} observe, large language models may hallucinate entities, infer non-existent attributes, or yield inconsistent outputs without domain-level guardrails. These factors highlight the need for reliable, schema-aware mechanisms that ensure essential policy elements are accurately extracted and verifiable. Finally, limited research \cite{ref12,ref17} focuses on validating generated policies and improving them based on linter feedback.

\subsection{Policy Externalization and AI Delegation}

Policy externalization has been a point of research interest with emphasis on modularity and verifiability. Xu and Zhang \cite{ref30} highlighted attribute-based access control for collaboration across multi-user and cross-organizational environments, where dynamic sharing requires fine-grained and context-aware decisions. Adeyinka \cite{ref2} showed that embedding authorization within applications limits scalability and auditability in hybrid systems, motivating externalized PaC frameworks. Ferraiolo et al. \cite{ref9} also emphasized that separating policies is valuable in distributed systems where authorization depends on dynamic attributes and context. Governance-as-a-service \cite{ref10} and work by Chopra \cite{ref6} extend this to multi-agent authorization, aligning with the system's conceptualization of the policy generator as a governed AI agent.

Building on this, South et al. \cite{ref27} proposed an authenticated delegation framework extending OAuth 2.0 and OpenID Connect for AI agents, enabling verifiable delegation from humans to agents. Their work introduces the concept of translating natural-language permissions into auditable access control configurations, aligning with the system's objective of bridging intent and machine-enforceable authorization.

\subsection{Security Failure Modes in LLM-Generated Policies}

We identify four recurring failure modes in direct LLM-generated access control policies that are not detectable through syntax validation or compilation alone.

\begin{itemize}
  \item \textbf{Deny-Rule Omission:} When a policy expresses denial intent, LLMs frequently emit \texttt{default allow := false} without an active \texttt{deny if} rule. The policy compiles and lints cleanly but enforces nothing and OPA evaluates the deny rule as undefined rather than triggered.
  \item \textbf{Constraint Incompleteness:} Conditional qualifiers such as ``for their team'' or ``when assigned'' are silently dropped, producing rules that are broader than the original intent. The generated Rego is syntactically complete but semantically weaker than the stated policy.
  \item \textbf{Over-Permissive Subject Matching:} References to a specific actor (e.g., ``the assigned nurse'') are translated to role-level predicates (e.g., \texttt{input.subject.role == "nurse"}), granting access to all role members rather than the designated individual.
  \item \textbf{Implicit Allow-All Paths:} Without \texttt{default allow := false}, allow rules return undefined rather than false for non-matching inputs. Depending on how the enforcement point handles undefined, this silently grants access.
\end{itemize}

The use of large language models for generating authorization policies introduces new security risks. Generated policies may be syntactically valid but semantically incorrect, leading to over-permissive access, missing constraints, or unintended privilege escalation. The solution mitigates some of these risks by enforcing structured decomposition, schema validation, and test-driven verification. However, incorrect extraction or incomplete schema definitions may still result in unintended policy behavior, highlighting the need for additional safeguards and human validation in security-critical environments.

Our evaluation quantifies these risks directly. A direct single-shot LLM baseline, given identical inputs, generates semantically incorrect deny Rego for 62.5\% of deny-decision policies, producing patterns such as \texttt{default allow := false} with no active deny rule, which fails to prohibit the targeted action. Additionally, 60\% of ownership-qualified policies and 44\% of status-qualified policies generated by the baseline are missing the corresponding Rego conditions, resulting in over-permissive rules that grant access to unauthorized subjects. The system's deterministic semantic safety guardrails reduce these failure modes to zero on the evaluated dataset.

\section{Design Rationale and Technical Choices}

This section explains the key design decisions that shape our approach, specifically the prompting strategy used for translating natural-language access control requirements into executable policies and the choice of policy language.

\subsection{Prompt Engineering for Policy Extraction}

A distinguishing aspect of the system is its structured prompting mechanism designed to address a main limitation of LLM-based policy generation: lack of control and reproducibility \cite{ref31,ref15,ref4}. While our modular prompting and schema, lint, and test guardrails reduce hallucinations and enforce structural validity, model inference can remain non-deterministic: repeated runs on the same NLACP may yield different extracted representations or different, yet valid, Rego implementations. The system therefore emphasizes deployability checks, including schema validation, linting, compilation, and unit tests, and auditability of intermediate artifacts rather than guaranteeing identical outputs across runs. This prompt chaining strategy, augmented by verification feedback, encourages more consistent and interpretable outputs by constraining each step to a fixed schema and by rejecting invalid generations. Few studies have explored prompt design as a formal control mechanism. Yang et al. \cite{ref31} and Lawal et al. \cite{ref15} used heuristic text-to-JSON prompts, whereas the system formalizes prompt orchestration into a schema-governed workflow with post-generation validation via Regal \cite{ref24} and OPA compilation and testing. In the system, prompt chaining supports:

\begin{itemize}
  \item \textbf{Intent identification:} specially crafted prompts based on few-shot prompting \cite{ref3} and program-of-thought prompting \cite{ref5} identify whether input represents a policy, simplify it, and extract multiple policies from provided input.
  \item \textbf{Element extraction:} prompts extract decision, subject, action, resource, condition, and purpose and return structured JSON for each recognized policy statement.
  \item \textbf{Rego synthesis:} prompts convert extracted elements into Rego policies.
\end{itemize}

\subsection{Rego as Policy Language}

Most recent research around translating NLACPs focuses on NGAC \cite{ref1,ref30,ref9,ref12} and XACML \cite{ref18,ref31,ref20,ref15,ref21}. For our pipeline, we chose Rego due to its declarative, human-readable syntax, enabling non-technical users to validate output and provide human-in-the-loop feedback. Rego is the policy language of Open Policy Agent \cite{ref22}, which is widely used and compatible with multiple projects \cite{ref23} such as Kubernetes, Envoy, Express, Terraform, and Linux-PAM. Given adoption of Rego, the system can be useful for a large set of practitioners.

\subsection{Baseline Comparison Design}
\label{sec:baseline}

To evaluate the contribution of structured decomposition, we compare the system against a direct single-shot LLM baseline (Arm A). The baseline receives the raw natural-language policy text and a fixed prompt:

\begin{verbatim}
Translate the following natural-language access control policy into
executable Rego code for Open Policy Agent (OPA).
Requirements:
- Generate valid Rego syntax with default-deny semantics.
- Return only the Rego module, no explanation.
- Preserve the meaning of the original policy as closely as possible.
Policy: "{policy text}"
\end{verbatim}

The baseline uses the same underlying LLM and temperature as the system's Rego generation step. It receives no DSARCP decomposition, no schema context, no linting feedback, and no semantic safety checks.

If the model returns no executable Rego output, the policy is treated as rejected. This behavior reflects a practical usage scenario where a single prompt is used without iterative refinement or structured decomposition.

This baseline represents the approach a practitioner would take prior to adopting a structured pipeline, and directly evaluates the question of whether explicit decomposition and validation improve reliability over direct prompting.

Both approaches are evaluated on identical inputs. Metrics are computed at the policy level; a policy is considered correct only if it simultaneously satisfies all four criteria: OPA compilation, Regal lint compliance, positive-test pass, and negative-test pass.

\section{System Architecture and Pipeline}

\begin{figure}[t]
  \centering
  \includegraphics[width=\textwidth]{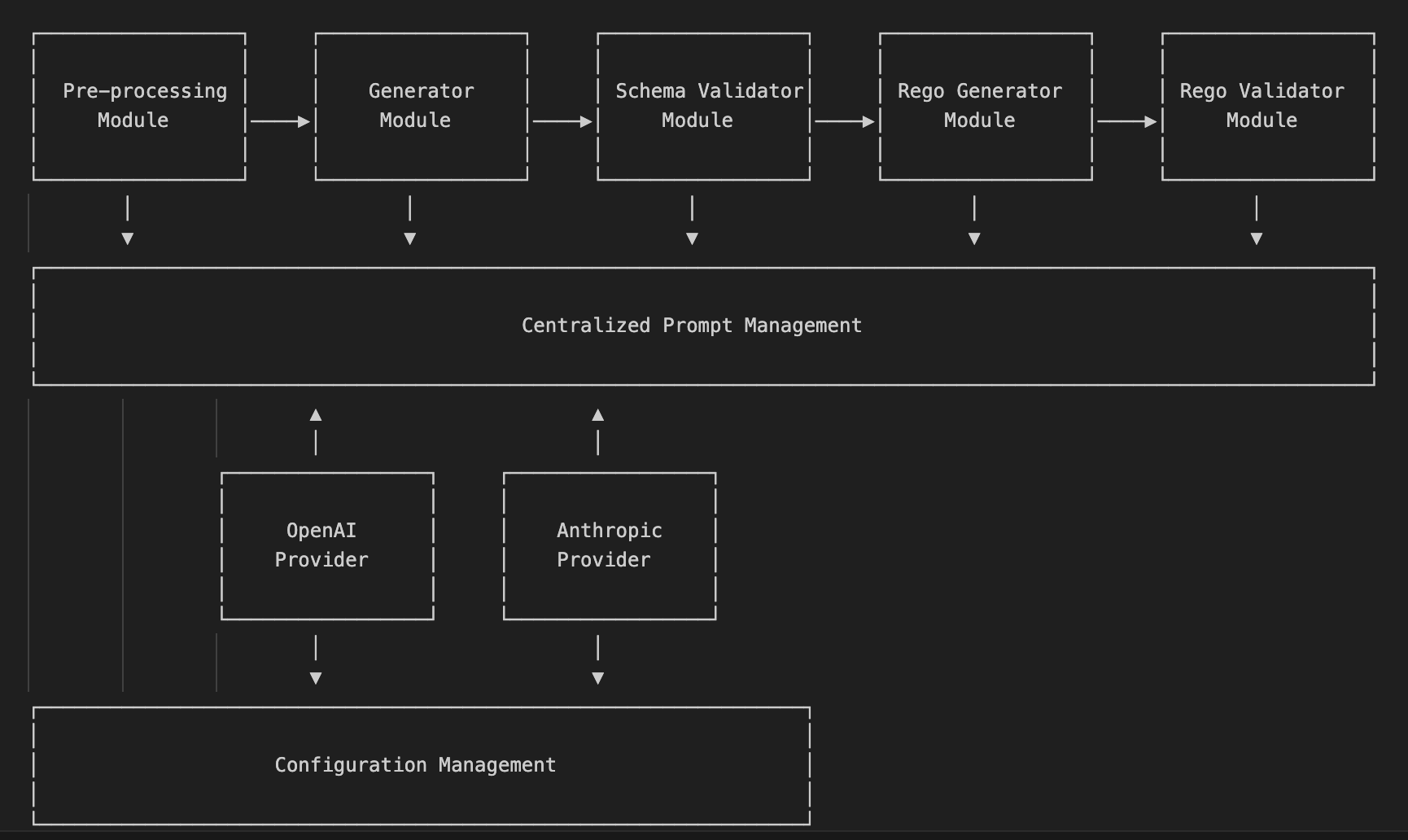}
  \caption{Solution architecture, showing the four core modules and supporting infrastructure.}
  \label{fig:architecture}
\end{figure}

The system is a modular and flexible tool chain that converts natural-language access control policies into executable Rego code. The pipeline is organized into four core modules: (a) pre-processing, (b) component extraction, (c) schema validation, and (d) Rego generation, refinement, and testing. These modules are supported by a centralized prompt management layer and a configurable LLM provider interface, as shown in Fig.~\ref{fig:architecture}. Each module can be invoked individually through the command line interface or orchestrated as a continuous flow through a user interface.

\subsection{Pre-processing}

The pre-processing module performs:

\begin{itemize}
  \item \textbf{Policy detection} by identifying if the statement represents one or multiple policies. The detection logic uses an explicit definition: an access control policy describes who, as the subject, can or cannot perform an action on a resource, sometimes under conditions or for a purpose. This filtering improves precision during pre-processing. Statements that lack an identifiable subject, action, or resource under this definition are treated as out-of-scope and flagged as non-policies.
  \item \textbf{Co-reference resolution} by resolving pronouns and implicit references.
  \item \textbf{Text segmentation} by breaking input text into multiple policy statements.
\end{itemize}

By combining co-reference resolution and text segmentation, an input such as ``Nurses are allowed to read prescriptions, but they are not allowed to change them'' becomes two NLACPs: ``Nurses are allowed to read prescriptions'' and ``Nurses are not allowed to change prescriptions.''

Unlike Text2Policy \cite{ref29} and ACRE \cite{ref26}, which rely on handcrafted syntactic patterns, the system couples prompt-based intent detection with deterministic normalization, providing a transparent pre-processing audit trail.

\subsection{Component Extraction}

The extraction module identifies core access control components using prompts to a large language model following few-shot mechanisms. The system extracts:

\begin{itemize}
  \item \textbf{Decision:} allow or deny determination, for example, ``allow,'' ``permit,'' or ``deny.''
  \item \textbf{Subject:} actor to whom the policy applies, for example, ``administrators'' or ``users.''
  \item \textbf{Action:} operation being regulated, for example, ``access,'' ``modify,'' or ``view.''
  \item \textbf{Resource:} protected asset, for example, ``database,'' ``files,'' or ``records.''
  \item \textbf{Condition:} contextual constraints, for example, ``during business hours'' or ``with approval.''
  \item \textbf{Purpose:} intended goal of the action, for example, ``for maintenance'' or ``for auditing.''
\end{itemize}

When the extraction step returns multiple components for a single policy statement, a pre-consolidation step merges components that share the same action, condition, and purpose but differ only in subject role or resource type. For example, ``nurses and doctors may view lab results, prescriptions, and appointments'' yields six components, which are consolidated into a single component with list-valued fields (e.g., \texttt{\{"subject": ["nurse","doctor"], "resource": ["lab result","prescription","appointment"]\}}).

The LLM therefore receives a single structured component instead of multiple redundant objects. This reduces prompt complexity and eliminates a primary source of redundant-block generation before the LLM is invoked.

This design directly contributes to improved lint pass rates and reduces structurally redundant policy blocks observed in baseline outputs.

\subsection{Schema Validation}

The schema validation module ensures extracted DSARCP components conform to predefined organization-specific schemas in JSON listing valid values for each component. If crucial components like subject, action, or resource are not identified, the flow stops and missing attribute details are shared with the policy administrator. This stage is configurable and can be disabled for research use. The system includes multiple schemas based on datasets used in prior research \cite{ref12}, which can be modified as needed.

Critically, the validated schema is also injected verbatim into the Rego generation system prompt, constraining the LLM to reference only field names and values declared in the organization's schema. This closes the gap between component-level validation and code-level attribute grounding. As a result, the generator cannot introduce undeclared fields such as \texttt{input.user.department}, even when the natural-language policy implies one. This constraint ensures consistency between extracted components and generated code, and prevents schema violations that would otherwise pass syntactic validation but fail at runtime.

\subsection{Rego Generation, Refinement, and Testing}

The generator emits Rego modules following a deny-by-default pattern and annotates rules with extracted DSARCP fields for auditability. If multiple policy statements are identified, the output is a single Rego module encoding all statements. Generated Rego passes through a three-stage validation and repair process before being returned.

\paragraph{Redundant-block consolidation.}
After initial generation, a deterministic post-processing step inspects the emitted Rego for allow or deny blocks that are identical except for a single field value. Such blocks are collapsed into a single rule using set membership notation (e.g., \texttt{input.subject.role in \{"nurse", "doctor"\}}). This step runs before the feedback loop and removes a common source of Regal messy-rule violations caused by LLM verbosity.

\paragraph{Structural validation feedback loop.}
The consolidated Rego is compiled using \texttt{opa check} and linted with Regal \cite{ref24}. Any compilation errors or lint findings are formatted as structured messages and submitted to the LLM together with the original Rego and extracted components for correction. This loop runs up to three iterations.

\paragraph{Semantic safety guardrails.}
Two post-generation checks operate against the original natural-language policy text outside the LLM inference path:

\begin{itemize}
  \item \textbf{Ownership guardrail:} If the policy contains an ownership qualifier such as ``their own records'' or ``his or her own prescriptions,'' but the generated Rego lacks a condition such as \texttt{input.resource.owner == input.subject.id}, a targeted error is injected into the feedback loop.
  \item \textbf{Status guardrail:} If the policy contains a status qualifier such as ``pending,'' ``upcoming,'' or ``in transit,'' but the generated Rego lacks a corresponding \texttt{input.resource.status} condition, the missing constraint is identified and fed back.
\end{itemize}

These guardrails reduce ownership-condition omission from 60\% to 0\% and status-condition omission from 44\% to 0\% on the evaluated dataset. The system prompt additionally prohibits the \texttt{allow if not deny} pattern for deny-decision policies and instead requires an explicit \texttt{deny if \{...\}} rule. This prevents the semantically incorrect pattern of emitting \texttt{default allow := false} without an active deny condition.

\paragraph{Test generation and repair.}
Unit tests are automatically generated from DSARCP components. The system parses each generated Rego rule to extract field-value constraints and merges any missing fields into the LLM-proposed positive test inputs so that all allow conditions are satisfied.

For negative tests, a repair step detects inputs that accidentally satisfy all allow conditions and deterministically breaks one restrictive condition. For policies using \texttt{not deny} semantics, the negative input is constructed to satisfy the deny rule rather than simply negating an allow condition.

\section{Solution Features}

The system provides a web-based interface that unifies the modular pipeline into an interactive environment. The design supports practitioners who author, validate, and deploy policies, and researchers who experiment with prompting, evaluation, and schema configuration.

\subsection{Understanding and Experimenting with the Flow}

The single-policy view serves as an exploratory environment for understanding how natural language statements evolve into executable Rego policies. A user can paste a statement and observe each stage from policy detection and DSARCP extraction to schema validation and code synthesis. The interface displays intermediate reasoning and explanations for transparency.

Practitioners can use this mode to fine-tune access rules and verify logic, while researchers can analyze prompt behavior and evaluate extraction accuracy. The same interface allows editing outputs generated by modules, enabling controlled experimentation with alternative phrasing, few-shot examples, or context windows.

\subsection{Batch Processing and Corpus Preparation}

The batch processing feature enables users to run the system across large text or groups of policy statements, following the configured flow for all statements. This is valuable for teams onboarding to policy-based access control or extracting policies from design documents, user stories, or functional requirement documents, as highlighted by Lawal et al. \cite{ref15}. Users can download generated Rego modules and import them into version control systems. For researchers, this tab supports experiments comparing prompting variants or model configurations across identical datasets.

\subsection{Testing and Validation of Generated Policies}

Once policies are generated, the testing and validation feature verifies correctness and deployment readiness. The system runs each policy through \texttt{opa check} and linting with Regal followed by unit tests executed using \texttt{opa test}. It generates positive and negative test cases derived from DSARCP components. This test suite bridges generation accuracy with runtime semantics, ensuring policies are both syntactically valid and operationally sound.

The system offers users the choice between LLM-based and rule-based test generation modes. While both modes are effective for standard DSARCP structures, test generation faces challenges with policies referencing external constants, nested or compound conditions, contextual variables, type conversions, or time-based evaluations. These limitations impact test generation logic rather than the correctness of generated Rego code. The LLM-based mode provides improved handling of some complex cases, but certain challenges persist in both modes.

To improve test reliability beyond what the LLM alone can provide, the system applies two deterministic repair steps. First, the body of each generated Rego rule is parsed to extract all constraint expressions of the form \texttt{input.X == "value"}, including set-membership and cross-field equality conditions. Any fields absent from the LLM-generated test input are then merged in so that positive test inputs satisfy every condition in the rule, not just the ones anticipated by the model.

Second, negative test inputs are checked for accidental satisfiability. If a proposed negative input satisfies all allow conditions, which can occur when the LLM assigns identical values to fields such as \texttt{resource.owner} and \texttt{subject.id}, one restrictive condition is broken deterministically. For policies using \texttt{not deny} semantics, the negative input is instead constructed to trigger the deny rule so that the allow rule cannot fire.

\subsection{Configuration and Prompt Customization}

The configuration tab provides control over how policies are extracted, validated, generated, and tested. Users can select among LLM providers based on cost, accuracy, or privacy requirements, can toggle schema validation and Rego validation, and can also decide if they want to use a large language model or rules to generate test cases. Users can also customize prompts by editing prompt templates to refine behavior for specific domains or datasets. This makes the system adaptable for production and extensible for research experimentation without modifying the core engine.

\section{Evaluation and Results}

The system was evaluated on the ACRE dataset \cite{ref26}, consisting of 485 access control statements used in prior work \cite{ref12,ref26}. After applying a subject-action-resource completeness filter to retain only statements with non-null values for all three required fields, 372 policies remained for evaluation.

Of the 372 ACRE-complete policies, the system accepts 206 as valid NLACPs and rejects the remaining 166 inputs, which lack a clearly identifiable subject, action, or resource under the pipeline's strict NLACP definition. All 372 policies are evaluated through both arms to ensure a consistent comparison against dataset-defined ground truth.

\subsection{Experimental Setup}

We compare two arms:

\begin{itemize}
  \item \textbf{Arm A --- Direct LLM baseline:} the raw natural-language policy is submitted with a single fixed prompt requesting valid Rego (Sect.~\ref{sec:baseline}). No decomposition, schema context, or validation feedback is provided. If the model returns no executable output, the policy is treated as rejected.
  \item \textbf{Arm B --- The system:} the full pipeline including pre-processing, DSARCP extraction, schema validation, Rego generation with structured validation, and automated testing.
\end{itemize}

Both arms use Claude (Anthropic) as the underlying LLM with temperature 0.0. This design isolates the impact of structured decomposition and validation by holding the model and input set constant.

A policy is considered fully correct only if it simultaneously satisfies all four automated criteria: OPA compilation, Regal lint compliance, positive-test pass, and negative-test pass. We define effective correct coverage as the fraction of ACRE-complete policies for which a fully correct policy is produced.

We note that this definition of correctness reflects operational validity under compilation, linting, and test-based verification rather than full semantic equivalence to the original policy intent.

\subsection{Overall Correctness}

Table~\ref{tab:overall} reports results across all 372 ACRE-complete policies.

\begin{table}[t]
\centering
\caption{Correctness across all 372 ACRE-complete policies.}
\label{tab:overall}
\begin{tabular}{lcc}
\toprule
\textbf{Metric} & \textbf{Direct LLM} & \textbf{Solution Pipeline} \\
\midrule
Input acceptance rate (attempted) & 20.7\% (77/372) & 55.4\% (206/372) \\
Compile OK & 20.7\% & 54.0\% \\
Lint pass & 17.5\% & 53.8\% \\
Positive test pass & 17.7\% & 51.3\% \\
Negative test pass & 19.9\% & 53.2\% \\
Effective correct coverage & 15.3\% & 50.3\% \\
\bottomrule
\end{tabular}
\end{table}

The direct LLM produces fully correct policies for fewer than one in six of ACRE-complete inputs, while the system achieves substantially higher effective coverage. The primary failure mode for the baseline is rejection, where valid access control requirements receive no executable policy output.

\subsection{Head-to-Head Comparison}

Table~\ref{tab:headtohead} examines the 77 policies where both arms generate Rego output, isolating generation quality from coverage effects.

\begin{table}[t]
\centering
\caption{Head-to-head comparison on 77 overlapping policies.}
\label{tab:headtohead}
\begin{tabular}{lcc}
\toprule
\textbf{Metric} & \textbf{Direct LLM} & \textbf{Solution Pipeline} \\
\midrule
Compile OK & 100.0\% & 97.4\% \\
Lint pass & 84.4\% & 96.1\% \\
Positive test pass & 85.7\% & 92.2\% \\
Negative test pass & 96.1\% & 97.4\% \\
All four pass & 74.0\% & 90.9\% \\
\bottomrule
\end{tabular}
\end{table}

The baseline's 100\% compile rate reflects selection bias, as it only attempts simpler policies. For the same inputs, the system generates richer policies that include ownership checks, status conditions, and consolidated rule structures. Despite slightly higher structural complexity, the system achieves higher lint compliance, test pass rates, and overall correctness.

\subsection{Security-Specific Metrics}

Table~\ref{tab:security} evaluates security-relevant properties on the overlap set.

\begin{table}[t]
\centering
\caption{Security-specific metrics (77-policy overlap set).}
\label{tab:security}
\begin{tabular}{lcc}
\toprule
\textbf{Property} & \textbf{Direct LLM} & \textbf{Solution Pipeline} \\
\midrule
Deny-rule completeness ($n=8$) & 37.5\% (3/8) & 87.5\% (7/8) \\
Role check present$^{\dagger}$ & 90.9\% & 87.0\% \\
Ownership condition ($n=5$) & 40.0\% & 100\% \\
Status condition ($n=9$) & 55.6\% & 100\% \\
Correct default semantics & 100.0\% & 97.4\% \\
Hallucinated input fields & 0.0\% & 0.0\% \\
\bottomrule
\end{tabular}

\vspace{0.4em}
\begin{minipage}{0.96\linewidth}\footnotesize
$^{\dagger}$The system frequently consolidates subject roles into set-membership expressions (\texttt{input.subject.role in \{"nurse","doctor"\}}), which regex-based role-check detection does not match. All the system's policies flagged as missing a role check pass positive and negative semantic tests, indicating the lower figure is a measurement artefact rather than a security gap.
\end{minipage}
\end{table}

Deny-rule completeness measures whether deny-decision policies generate an active \texttt{deny if} rule. The baseline frequently produces patterns such as \texttt{default allow := false} without an explicit deny rule, which fails to enforce the intended restriction. This defect is not detectable through compilation or linting and may only surface at runtime.

Ownership and status condition checks measure whether policies containing these qualifiers produce corresponding conditions such as \texttt{input.resource.owner} or \texttt{input.resource.status}. Missing these conditions results in over-permissive policies. The system significantly reduces these failure modes on the evaluated dataset.

The role-check gap reflects a measurement limitation. The system often uses set membership expressions (e.g., \texttt{in \{...\}}), which are not detected by simple pattern matching but remain semantically correct and pass all tests.

\subsection{Discussion}

The results show that structured decomposition plays a critical role in improving both coverage and correctness. While the baseline produces reasonable outputs when it attempts a policy, it leaves a large fraction of access control requirements without machine-enforceable representation.

The lower positive-test pass rate relative to the negative-test pass rate (51.3\% vs. 53.2\%) reflects the structural asymmetry between the two test types: positive tests require all extracted conditions to be simultaneously satisfied, while negative tests only require one condition to be violated, making them inherently easier to construct and pass.

Security-specific failures highlight a deeper issue. Deny-rule omissions produce policies that appear correct but fail under critical conditions. These defects are not visible through static checks and represent a meaningful risk in practical deployments.

\section{Conclusion and Future Work}

The system democratizes the translation of natural-language access requirements into executable, auditable policies. Our controlled comparison against a direct LLM baseline shows that structured decomposition and schema-aware validation play a critical role in producing security-correct policies. The baseline leaves a large fraction of ACRE-complete policies without executable output and generates semantically incorrect deny rules for a substantial portion of the deny policies it attempts. Its modular pipeline enables both precision and adaptability in enterprise environments while allowing researchers to improve the performance and accuracy of the translation process. In uniting these two audiences, the system functions as both a policy-as-code workbench and a research testbed. It lays a foundation for reproducible experiments and practical deployment.

Future enhancements will focus on improving adaptability, scalability, and reliability:

\begin{itemize}
  \item \textbf{Advanced schema support:} We plan to enhance schema validation to support wildcards, pattern-based matching, and hierarchical structures, making the tool more scalable for large or dynamic domains and reducing the need for manual enumeration of values.
  \item \textbf{Independent test generation:} To further increase trust in automated validation, we aim to decouple policy and test generation, potentially by using separate large language models or human-in-the-loop processes, to reduce the risk of correlated errors.
  \item \textbf{Enhanced support for complex policies:} We will improve the robustness of test generation and policy synthesis for policies with nested conditions, contextual variables, or references to external constants.
  \item \textbf{Multi-language output:} We plan to extend support beyond Rego to other policy languages such as Cedar, XACML, and NGAC, enabling broader adoption across diverse access control ecosystems.
  \item \textbf{Extended semantic safety coverage:} The current guardrails address ownership and status qualifiers. We plan to extend coverage to temporal conditions (``during business hours''), hierarchical role inheritance, and purpose-based constraints, building toward a comprehensive semantic safety specification.
\end{itemize}

\paragraph{Disclosure of Interests.}
The authors have no competing interests to declare that are relevant to the content of this article.

\appendix
\section{Additional Resources and Details}

This appendix provides supplementary material intended to support reproducibility and does not introduce new technical contributions.

\subsection{Example NLACP to Rego Translation}

The diagram shows how a policy statement goes through the different modules, along with the inputs and outputs for each module.

\begin{figure}[p]
  \centering
  \includegraphics[width=\textwidth]{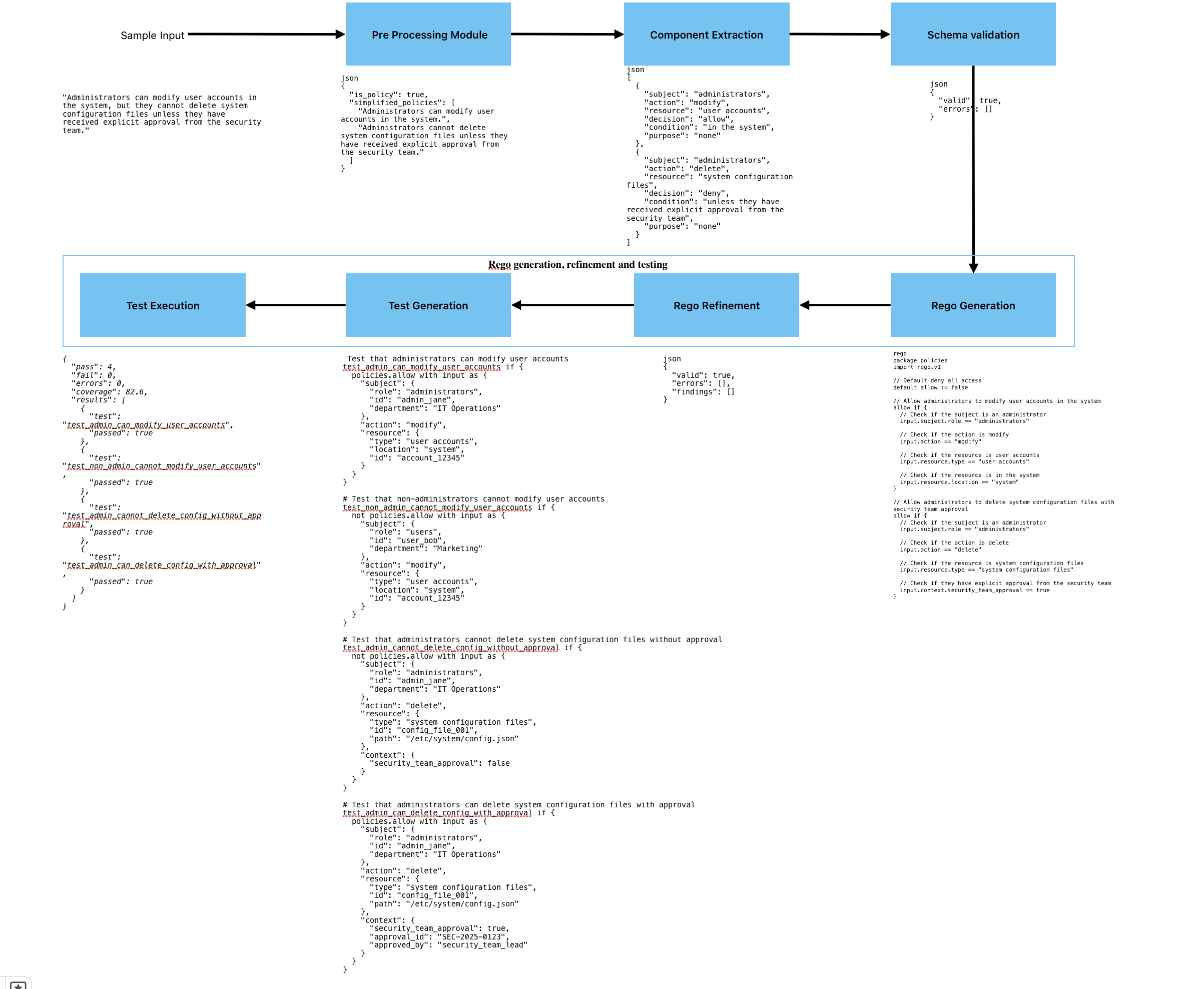}
  \caption{Inputs and outputs of each component.}
  \label{fig:example}
\end{figure}

\noindent\textbf{Note:} Code for the tool described in this paper will be made publicly available upon acceptance.


\begin{thebibliography}{32}

\bibitem{ref1}
M. Abdelgawad, I. Ray, and S. Alqurashi, ``Synthesizing and analyzing ABAC models from natural language policies,'' in \emph{Proceedings of the ACM Symposium on Access Control Models and Technologies (SACMAT '23)}, 2023.

\bibitem{ref2}
A. Adeyinka, ``Automated compliance management in hybrid cloud architectures: A policy-as-code approach,'' ResearchGate, 2023.

\bibitem{ref3}
T. B. Brown, B. Mann, N. Ryder, et al., ``Language models are few-shot learners,'' in \emph{Advances in Neural Information Processing Systems (NeurIPS 2020)}, 2020. \url{https://doi.org/10.48550/arXiv.2005.14165}

\bibitem{ref4}
M. Chen, J. Tworek, H. Jun, et al., ``Evaluating large language models trained on code,'' \emph{arXiv preprint arXiv:2107.03374}, 2021. \url{https://doi.org/10.48550/arXiv.2107.03374}

\bibitem{ref5}
M. Chen, D. Zhou, N. Sch\"arli, and L. Hou, ``Program of thoughts prompting: Disentangling computation from reasoning for numerical reasoning tasks,'' \emph{arXiv preprint arXiv:2305.20050}, 2023. \url{https://doi.org/10.48550/arXiv.2305.20050}

\bibitem{ref6}
V. Chopra, ``OAuth is not enough: Authorization challenges for autonomous AI agents,'' TechRxiv, 2025.

\bibitem{ref7}
N. Chowdhary, T. Dutta, and S. Chattopadhyay, ``AutoPAC: Exploring LLMs for automating policy to code conversion in business organizations,'' in \emph{IEEE Conference Proceedings}, 2025.

\bibitem{ref8}
J. W. Cutler, C. Disselkoen, A. Eline, et al., ``Cedar: A new language for expressive, fast, safe, and analyzable authorization (extended version),'' 2024. \url{https://doi.org/10.48550/arXiv.2403.04651}

\bibitem{ref9}
D. F. Ferraiolo, R. Sandhu, and D. R. Ferraiolo, ``The policy machine: A novel architecture and framework for access control policy specification and enforcement,'' \emph{Journal of Systems and Software}, vol. 89, pp. 1--17, 2016.

\bibitem{ref10}
S. Gaurav, J. Heikkonen, and J. Chaudhary, ``Governance-as-a-service: A multi-agent framework for AI system compliance and policy enforcement,'' \emph{arXiv preprint arXiv:2508.18765}, 2025. \url{https://doi.org/10.48550/arXiv.2508.18765}

\bibitem{ref11}
S. H. Jayasundara, N. A. G. Arachchilage, and G. Russello, ``SoK: Access control policy generation from high-level natural language requirements,'' \emph{arXiv preprint arXiv:2310.03292}, 2023. \url{https://doi.org/10.48550/arXiv.2310.03292}

\bibitem{ref12}
S. H. Jayasundara, N. A. G. Arachchilage, and G. Russello, ``RAgent: Retrieval-based access control policy generation,'' \emph{arXiv preprint arXiv:2409.07489}, 2024. \url{https://doi.org/10.48550/arXiv.2409.07489}

\bibitem{ref13}
R. N. H. Khan, D. Wasif, J. H. Cho, and A. Butt, ``Multi-agent code-orchestrated generation for reliable infrastructure-as-code,'' \emph{arXiv preprint arXiv:2510.03902}, 2025. \url{https://doi.org/10.48550/arXiv.2510.03902}

\bibitem{ref14}
V. Kumar, W. Alasmary, and M. M. Hossain, ``Zero trust access control: A survey and research directions,'' \emph{IEEE Access}, 2024.

\bibitem{ref15}
S. Lawal, X. Zhao, A. Rios, R. Krishnan, and D. Ferraiolo, ``Translating natural language specifications into access control policies by leveraging large language models,'' in \emph{IEEE Conference on Trust, Privacy and Security in Intelligent Systems and Applications (TPS '24)}, 2024.

\bibitem{ref16}
K. Madan, ``Argen: Auto-regulation of generative AI via GRPO and policy-as-code,'' \emph{arXiv preprint arXiv:2509.07006}, 2025. \url{https://doi.org/10.48550/arXiv.2509.07006}

\bibitem{ref17}
A. Mittal and V. Venkatesan, ``Practical integration of large language models into enterprise CI/CD pipelines for security policy validation,'' \emph{IEEE Access}, 2025.

\bibitem{ref18}
M. Narouei, H. Khanpour, H. Takabi, et al., ``Towards a top-down policy engineering framework for ABAC,'' in \emph{Proceedings of the ACM Symposium on Access Control Models and Technologies (SACMAT '17)}, 2017.

\bibitem{ref19}
M. Narouei and H. Takabi, ``Automatic top-down role engineering framework using natural language processing techniques,'' in \emph{IFIP International Conference on Information Security Theory and Practice}, pp. 137--152, Springer, 2015.

\bibitem{ref20}
M. Narouei, H. Takabi, and R. Nielsen, ``Automatic extraction of access control policies from natural language documents,'' \emph{IEEE Transactions on Dependable and Secure Computing}, vol. 17, no. 3, pp. 506--517, 2018.

\bibitem{ref21}
OASIS Standard, ``eXtensible Access Control Markup Language (XACML) Version 3.0 Core Specification,'' 2013. \url{https://docs.oasis-open.org/xacml/3.0/xacml-3.0-core-spec-os-en.html}

\bibitem{ref22}
Open Policy Agent Contributors, ``Open Policy Agent,'' 2016. \url{https://www.openpolicyagent.org/}

\bibitem{ref23}
Open Policy Agent Contributors, ``OPA ecosystem REST API integrations,'' 2023. \url{https://www.openpolicyagent.org/ecosystem/rest-api-integration/}

\bibitem{ref24}
Open Policy Agent Ecosystem, ``Regal: Rego policy linter,'' 2024. \url{https://www.openpolicyagent.org/ecosystem/entry/regal}

\bibitem{ref25}
F. Romeo, L. Arena, F. Blefari, and F. A. Pironti, ``ARPAccino: An agentic-RAG for policy as code compliance,'' Springer, 2025.

\bibitem{ref26}
J. Slankas and L. Williams, ``Access control policy extraction from unconstrained natural language text,'' in \emph{International Conference on Social Computing}, 2013.

\bibitem{ref27}
T. South, S. Marro, T. Hardjono, et al., ``Authenticated delegation and authorized AI agents,'' \emph{arXiv preprint arXiv:2501.09674}, 2025. \url{https://doi.org/10.48550/arXiv.2501.09674}

\bibitem{ref28}
P. Subramaniam and S. Krishnan, ``Intent-based access control: Using LLMs to intelligently manage access control,'' \emph{arXiv preprint arXiv:2402.07332}, 2024. \url{https://doi.org/10.48550/arXiv.2402.07332}

\bibitem{ref29}
X. Xiao, A. Paradkar, S. Thummalapenta, and T. Xie, ``Automated extraction of security policies from natural-language software documents,'' in \emph{Proceedings of the ACM SIGSOFT Symposium on the Foundations of Software Engineering (FSE '12)}, 2012.

\bibitem{ref30}
D. Xu and Y. Zhang, ``Specification and analysis of attribute-based access control policies: An overview,'' in \emph{IEEE SERE-C}, 2014.

\bibitem{ref31}
M. Yang, V. Atluri, S. Sural, and A. Kundu, ``Extraction of machine enforceable ABAC policies from natural language text using LLM knowledge distillation,'' in \emph{Proceedings of the ACM Symposium on Access Control Models and Technologies (SACMAT '25)}, 2025. \url{https://doi.org/10.1145/3734436.3734447}

\bibitem{ref32}
T. Zhang, S. Pan, Z. Xing, and X. Sun, ``Deployability-centric infrastructure-as-code generation: An LLM-based iterative framework,'' \emph{arXiv preprint arXiv:2506.05623}, 2025. \url{https://doi.org/10.48550/arXiv.2506.05623}

\end{thebibliography}
\end{document}